\documentclass[conference]{IEEEtran}
\usepackage[latin9]{inputenc}
\usepackage{array}
\usepackage{multirow}
\usepackage{amsmath}
\usepackage{amssymb}
\usepackage{graphicx}

\makeatletter

\providecommand{\tabularnewline}{\\}

\IEEEoverridecommandlockouts
\usepackage{cite}
\usepackage{amsfonts}\usepackage{algorithmic}
\usepackage{textcomp}
\usepackage{xcolor}
\def\BibTeX{{\rm B\kern-.05em{\sc i\kern-.025em b}\kern-.08em
    T\kern-.1667em\lower.7ex\hbox{E}\kern-.125emX}}

\makeatother

\begin{document}
\title{Deep Clustering using Dirichlet Process Gaussian Mixture and Alpha
Jensen-Shannon Divergence Clustering Loss}
\author{\IEEEauthorblockN{Kart-Leong Lim} \IEEEauthorblockA{\textit{Institute of Microelectronics, A{*}Star} \\
 Singapore \\
 lkartl@yahoo.com.sg}}
\maketitle
\begin{abstract}
Deep clustering is an emerging topic in deep learning where traditional
clustering is performed in deep learning feature space. However, clustering
and deep learning are often mutually exclusive. In the autoencoder
based deep clustering, the challenge is how to jointly optimize both
clustering and dimension reduction together, so that the weights in
the hidden layers are not only guided by reconstruction loss, but
also by a loss function associated with clustering. The current state-of-the-art
has two fundamental flaws. First, they rely on the mathematical convenience
of Kullback-Leibler divergence for the clustering loss function but
the former is asymmetric. Secondly, they assume the prior knowledge
on the number of clusters is always available for their dataset of
interest. This paper tries to improve on these problems. In the first
problem, we use a Jensen-Shannon divergence to overcome the asymmetric
issue, specifically using a closed form variant. Next, we introduce
an infinite cluster representation using Dirichlet process Gaussian
mixture model for joint clustering and model selection in the latent
space which we called deep model selection. The number of clusters
in the latent space are not fixed but instead vary accordingly as
they gradually approach the optimal number during training. Thus,
prior knowledge is not required. We evaluate our proposed deep model
selection method with traditional model selection on large class number
datasets such as MIT67 and CIFAR100 and also compare with both traditional
variational Bayes model and deep clustering method with convincing
results. 
\end{abstract}

\section{Introduction}

One of the key success of deep learning is attributed to the feature
extraction capability that is found in the first several hidden layers
of the network. Deep clustering is an emerging topic in deep learning
where traditional clustering is performed in deep learning feature
space. However, clustering and deep learning are often mutually exclusive.
Deep learning is seen as a black box and the non-linearity of the
network remains mathematically unclear to the researcher. Traditional
clustering method such as Kmeans or GMM while mathematically sound
is very hard to extend to deep layer representation. A recent technique
known as deep clustering tries to bridge both sides by the simple
use of a minimization between the network weights and clustering parameters.
According to literature \cite{min2018survey}, we can categorize deep
clustering by their network architecture: autoencoder \cite{song2013auto,xie2016unsupervised,jiang2017variational,yang2019deep,yang2019deep2},
generative adversarial network \cite{zhou2018deep,ghasedi2019balanced}
and convolutional neural network \cite{hsu2017cnn,caron2018deep}.

In particular, we only focus on the autoencoder category which is
the simplest but performance wise is not necessarily inferior to the
other two categories \cite{min2018survey}. Within this category,
the Type II regularizer in Table I (also referred to as the clustering
loss in \cite{min2018survey}) is the basis for deep clustering. An
autoencoder (AE) maps an input to a point in a reduced dimension latent
space using reconstruction loss. Dimension reduction is important
for image clustering since a more compact representation can usually
improve the accuracy. Specifically, Autoencoder Based Clustering (ABC)
\cite{song2013auto,yang2017towards} proposed a Type II regularizer
to minimize the difference between an image encoded as a point in
the latent space and its nearest cluster mean. The latter is computed
using Kmeans in the entire latent space. When backpropagating from
this regularizer, the encoder weights will restructure itself to output
a latent space that mimics a partition performed by Kmeans. We can
achieve better clustering this way than either Kmeans or autoencoder
alone. In such approach, there are two objectives to be solved, i.e.
the network weights and the cluster parameters. Thus in practice,
we alternate between each optimization while fixing the other.

Along the way, self-expressiveness and sparse coding were proposed
on top of ABC's loss \cite{ji2017deep,sun2018learning}. Later on,
Gaussian mixture model (GMM) and Variational Autoencoder (VAE) quickly
became very popular for deep clustering \cite{dilokthanakul2016deep}
but remains unclear. The latest installment in this direction \cite{yang2019deep,jiang2017variational,yang2019vsb,lim2020}
uses the Kullback-Leibler divergence (KLD) between VAE and GMM to
define the Type II regularizer. Unlike the ABC's Type II regularizer
which is a point estimate version, the KLD approach is a powerful
representation since the performance can vary depending on the probabilistic
model used. Yet there are two main issues that are left out with the
latest installment of KLD based methods.

$\;$

\textbf{Problem} \textbf{statement}: 
\begin{verse}
1) KLD is asymmetrical and undefined when one of the probability distribution
is zero.

2) The number of cluster is a prior knowledge that is required beforehand
in deep clustering. 
\end{verse}
$\;$

The first issue will be problematic when there are regions in the
latent space that are left out by the network when they actually fall
under GMM coverage. This will lead to no gradient learnt for these
specific region. The second issue is that deep clustering are mostly
reported on trival cases such as 4 or 10 classes. When presented with
an unknown dataset, there is no easy way to know what is the optimal
number of clusters to set for deep clustering. Even if we resort to
well known technique like cross validation or Bayesian information
criteria, these methods are standalone techniques and do not jointly
optimize deep clustering. Also, the application of deep clustering
will be severely limited as a wrong decision on the number of clusters
will lead to much worse result than traditional clustering alone. 

$\;$

\textbf{Contributions}: 
\begin{verse}
1) We use Jensen-Shannon divergence for the loss function since it
is symmetrical and always defined.

2) Dirichlet process Guassian mixture is used to jointly perform both
deep clustering and model selection.
\end{verse}
$\;$

There is no closed form solution for the Jensen-Shannon divergence
(JSD) of two Gaussians. As a workaround, we use a recent work known
as the $\alpha$JSD in \cite{nielsen2019jensen} which rely on an
input skew parameter $\alpha$ to allow JSD to derive a closed form
solution. We also discuss a simpler form of $\alpha$JSD which is
essentially a first order solution i.e. cluster mean only. We propose
using Dirichlet process mixture (DPM) \cite{blei2017variational}
which is a type of Bayesian nonparametic model for clustering. The
main difference between DPM and GMM lies in the infinite number of
cluster representation. This means that we do not need to specify
exactly the number of clusters to be used. Instead, we fix a sufficiently
large truncated value for the infinite number of clusters. Also, the
Bayesian approach to DPM allows prior assumption over the latent space.
Thus, data overfitting and cluster singularities are simultaneously
addressed \cite{bishop2006pattern}. This is not possible with the
standard GMM in \cite{yang2019deep,jiang2017variational}.

\subsection{Related work}

\begin{table*}
\caption{Types of loss functions used in autoencoder}

\begin{centering}
\begin{tabular}{|c|c|c|c|c|}
\hline 
Objective & Methods & Reconst. Loss & Type I Regularizer Loss & Type II Regularizer Loss\tabularnewline
\hline 
\multirow{3}{*}{RC} &  &  &  & \tabularnewline
 & AE \cite{hinton2006reducing} & $p(x|z)$ & - & -\tabularnewline
 &  &  &  & \tabularnewline
\hline 
 &  &  &  & \tabularnewline
RC+PR & VAE \cite{kingma2014stochastic} & $p(x|z)$ & $KLD\left[q(z\mid x)\parallel p(z)\right]$ & -\tabularnewline
 &  &  &  & \tabularnewline
\hline 
\multirow{3}{*}{RC+PR+CL} &  &  &  & \tabularnewline
 & VAED \cite{lim2020} & $p(x|z)$ & $KLD\left[q(z\mid x)\parallel p(z)\right]$ & $KLD\left[q(z\mid x)\parallel p(z\mid\theta^{*})\right]$\tabularnewline
 &  &  &  & \tabularnewline
\hline 
\multirow{3}{*}{RC+CL+MS} & \multirow{3}{*}{This work} & \multirow{3}{*}{$p(x|z)$} & \multirow{3}{*}{-} & \multirow{3}{*}{$JSD\left[q(z\mid x)\parallel p(z\mid\phi^{*})\right]$}\tabularnewline
 &  &  &  & \tabularnewline
 &  &  &  & \tabularnewline
\hline 
\end{tabular}
\par\end{centering}
(RC, PR, CL, MS refers to reconstruction, prior representation, clustering
and model selection respectively)
\end{table*}

Table I summarizes some recent works related to this paper, notably
AE, VAE and VAED. Type I regularizer enforces the latent space of
the autoencoder to follow a prior distribution. Thus, we can generate
a sample in the latent space using a random number generator (associated
with the prior). The Type II regularizer minimizes the within class
distance in the latent space based on a classical clustering algorithm
such as Kmeans. Thus, samples coming from the same cluster will be
densely distributed in the latent space, allowing more discriminative
class pattern to be picked up later. Both Type I and II regularizers
are mutually exclusive and can be trained separately. In this work,
we mainly focus on the Type II regularizer. The works of VaDE and
VAED \cite{jiang2017variational,lim2020} are based on the KLD representation
of ABC \cite{song2013auto} while DGG \cite{yang2019deep} based on
the JSD representation of ABC. However, since there is no closed form
approach to JSD, the authors of DGG proposed a heuristic approach
to JSD. DPM has also been proposed by Ye et al \cite{ye2018nonparametric}
for VAE. However, the problem Ye et al is solving is subspace clustering
which is different from deep model selection. Their goal is to automatically
find the number of subspaces using DPM. Whereas, we are finding the
number of clusters to represent the entire latent space. In VaDE,
the underlying prior distribution is based on the original VAE i.e.
Gaussian distributed. In SB-VAE \cite{nalisnick2017stick}, the authors
model VAE's Type I regularizer with a Dirichlet process prior. In
VSC-DVM \cite{yang2019vsb}, the authors proposed using SB-VAE's Type
I regularizer and VaDE's Type II regularizer. Objectively, the authors
of VSC-DVM were solving finite cluster representation in the latent
space. To the best of our knowledge, it is uncommon to find model
selection in Type II regularizer.

\section{Methodology}

\subsection{Infinite number of clusters for representing latent space}

We begin with the simplest Type II regularizer known as the\textbf{
}ABC loss in \cite{song2013auto}. An autoencoder (AE) learns to represent
an image in the reduced dimension latent space by minimizing the reconstruction
loss (using the mean square error, MSE) between a input $x$ and network
output $y$ as seen in the first term of eqn (1).

\begin{equation}
\begin{array}{c}
\mathcal{L_{MSE}}+\mathcal{L_{ABC}}\\
\\
=\ln p(x|z)-\lambda_{3}\left\Vert \eta^{*}-z\right\Vert ^{2}\\
\\
=-\frac{1}{2}\left(x-y\right)^{2}-\frac{\lambda_{3}}{2}\left(\eta^{*}-z\right)^{2}
\end{array}
\end{equation}

A simple yet effective way to optimize both dimensionality reduction
and clustering is to introduce a second term in eqn (1) as proposed
recently by \cite{song2013auto}. The second term or the Type II regularizer
tries to enforce the latent space to exhibit a Kmeans like partitioning.
More specifically in eqn (1): Given an input image $x_{n}$, we use
the encoder to obtain a point, $z_{n}$ in the latent space. We first
use Kmeans to partition the latent space into $K$ clusters. Then,
we use the cluster assignment of Kmeans to find the nearest cluster
mean $\eta^{*}$ to the point $z_{n}$ in the latent space. Backpropagating
on this regularizer will affect the encoder weights $w$ to output
a re-structured latent space that mimic a partitioning performed by
Kmeans. The parameter $\lambda_{3}$ reduces the effect of the regularizer
on the encoder weights updates. The parameter $\lambda_{3}$ is set
to $[0,1]$.

In the above ABC approach, we are required to predefine the number
of clusters or $K$ to represent the latent space of the autoencoder,
whose mean are defined as \textbf{$\eta=\left\{ \eta{}_{k}\right\} _{k=1}^{K}\in\mathbb{R}{}^{D}$}
(The total dimension of each observed instance is denoted $D$). Instead
of predefining $K$, we would like to automatically find $K$. This
is especially helpful when we do not have label information for a
given dataset we wish to perform ABC. First, we represent the same
latent space using an infinite number of clusters $\eta=\left\{ \eta{}_{k}\right\} _{k=1}^{\infty}\in\mathbb{R}{}^{D}$.
Ideally, we would like to have a handful of dominating clusters that
carry larger weightage of information, $v=\left\{ v_{k}\right\} _{k=1}^{\infty}\in\mathbb{R}$
to represent the entire latent space. As we gradually approach the
other end of the spectrum, we would like to see a lot of redundant
clusters with lower weightage of information. These redundant clusters
will be slowly pruned away by optimization. When we plot the frequency
of this spectrum, we should observe a decaying distribution best modeled
by a Dirichlet distribution or $v\sim Beta(1,a_{0})$. We can normalize
the weightage using $\pi_{k}=v_{k}\prod_{l=1}^{k-1}\left(1-v_{k}\right)$
such that the sum of these normalized weightage across the spectrum
gives a unit measure i.e. $\sum_{k=1}^{\infty}\pi_{k}=1$ as given
by \cite{Sethuraman1994}. This autoencoder latent space representation
of using an infinite number of clusters is one of the key concept
in this paper.

\begin{figure*}
\begin{centering}
\includegraphics[scale=0.7]{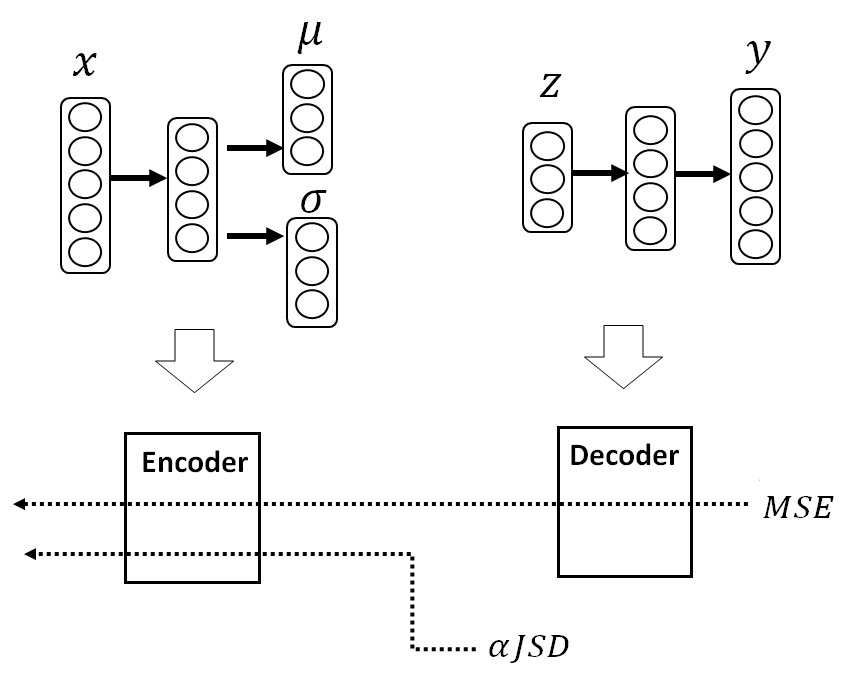}
\par\end{centering}
\caption{Deep model selection using proposed loss function $\alpha JSD$.}
\end{figure*}

\subsection{Regularizing autoencoder with infinite number of clusters}

We refer to the Type II regularizer used by deep clustering methods
e.g. \cite{jiang2017variational,yang2019vsb,lim2020} in eqn (2).
$q(z\mid x)$ is a single point in the VAE latent space which follows
a Gaussian distribution and $p(z\mid\theta^{*})$ is the GMM estimated
on the entire VAE latent space. A random sample in the VAE latent
space is defined by Kingma et al \cite{kingma2014stochastic} as $z=\mu+\sigma\cdot\mathcal{N}(0,1).$
Both $\mu$ and $\sigma$ are the hidden layers of an autoencoder.
To recover an autoencoder from VAE, we simply discard $\sigma$. On
the other hand, we define $\theta=\left\{ \eta,\tau,\psi\right\} $
as the GMM mean, precision and soft assignment. The cluster assignment
is denoted $\varsigma=\left\{ \varsigma_{n}\right\} _{n=1}^{N}$ where
$\varsigma_{n}$ is a $1-of-K$ binary vector, subjected to $\sum_{k=1}^{K}\varsigma_{nk}=1$
and $\varsigma_{nk}\in\left\{ 0,1\right\} $. $\theta^{*}$ refers
to the parameters of the optimal GMM cluster, given a particular sample
in the latent space, $z_{n}$. It is found by the cluster assignment
$\varsigma_{n}$, using eqn (2).

\begin{equation}
\begin{array}{c}
{KLD}\left[q(z\mid x)\parallel p(z\mid\theta^{*})\right]\\
\\
\theta^{*}=\underset{k}{\arg\max}\;\left\{ \ln\mathcal{N}\left(z_{n}\mid\eta_{k},\tau_{k}\right)^{\varsigma_{nk}}+\left(\ln\psi_{k}\right)^{\varsigma_{nk}}\right\} 
\end{array}
\end{equation}
\\

In most VAE based deep clustering e.g. \cite{yang2019vsb}, the authors
assume $\psi_{k}=\frac{1}{K}$ i.e. each cluster has equal sample
count. Thus, the second term in $\theta^{*}$ can be discarded when
computing the regularizer in eqn (2).

Next, we extend the Type II regularizer to DPM as denoted by $p(z\mid\phi^{*})$
in eqn (3) and also illustrated in Figure 1. We further define $\phi=\left\{ \eta,\tau,v\right\} $
as the DPM mean, precision, and cluster weightage. The cluster assignment
is also denoted $\varsigma$. The meaning of $\phi^{*}$ is similar
to $\theta^{*}$ where the optimal DPM parameters is found by solving
the DPM cluster assignment in eqn (3).

\begin{equation}
\begin{array}{c}
{KLD}\left[q(z\mid x)\parallel p(z\mid\phi^{*})\right]\\
\\
\phi^{*}=\underset{k}{\arg\max}\left\{ \underset{\eta_{k},\tau_{k},v_{k}}{E}\left[\ln\mathcal{N}(z_{n}\mid\eta_{k},\tau_{k})^{\varsigma_{nk}}\right.\right.\\
\\
\left.\left.+\left(\ln v_{k}\right)^{\varsigma_{nk}}+\sum_{l=1}^{k-1}\left(\;\ln(1-v_{l})\;\right)^{\varsigma_{nk}}\right]\right\} 
\end{array}
\end{equation}
\\
One of the main difference between deep clustering in (2) and deep
model selection in (3) can be attributed to the difference between
$\psi_{k}=\frac{1}{K}$ in eqn (2) and $\pi_{k}=v_{k}\prod_{l=1}^{k-1}(1-v_{l})$
in eqn (3). The cluster size of the former is $1\ldots k\ldots K$
and the latter is $1\ldots k\ldots\infty$. Moreover, in the graphical
representation between GMM and DPM in Figure 3, we can see that the
DPM parameters in deep model selection rely on prior assumptions and
are in turn governed by several hyperparameters. In Fig 3, $\omega_{0}$
is the hyperparameter of the Beta distributed prior for cluster weightage.
Likewise, $\lambda_{0}$ and $m_{0}$ correspond to the Gaussian distributed
prior for cluster mean and $a_{0},b_{0}$ are for the Gamma distributed
prior for precision.

\subsection{Type II Regularizer using $\alpha$JS divergence}

It is recently shown in \cite{lim2020} that when we re-express the
ABC objective as a KLD between GMM and VAE encoder output i.e. between
two Gaussians, a generic closed form solution is available \\
 
\begin{equation}
\begin{array}{c}
{KLD}\left[q(z\mid x)\parallel p(z\mid\theta^{*})\right]\\
\\
={KLD}\left[\mathcal{N}(\mu,\sigma)\parallel\mathcal{N}(\eta^{*},\tau^{*})\right]\\
\\
=\ln\tau^{*}+\ln\sigma+\frac{\left(\tau^{*}\right)^{-1}+\left(\eta^{*}-\mu\right)^{2}}{2\sigma_{2}}-\frac{1}{2}
\end{array}
\end{equation}
\\
Unlike the KLD of two Gaussians, there is no closed-form solution
for JSD. The authors in \cite{yang2019deep} propose a JSD for the
ABC objective based on an upper bound constraint. The former is essentially
an average sum between two standard KLD of eqn (2). Instead, we consider
another form of JSD which we called the $\alpha$JSD: In Nielsen \cite{nielsen2019jensen},
the use of a skew parameter $\alpha$ allows the JSD to possess a
closed form solution between two Gaussians as quoted in eqn (5)

\begin{equation}
\begin{array}{c}
{\alpha JSD}\left[\mathcal{N}(\mu_{1},\sigma_{1})\parallel\mathcal{N}(\mu_{2},\sigma_{2})\right]\\
\\
=\left(1-\alpha\right)\left(\mu_{\alpha}-\mu_{1}\right)^{2}\sigma_{\alpha}^{-1}+\alpha\left(\mu_{\alpha}-\mu_{2}\right)^{2}\sigma_{\alpha}^{-1}\\
\\
-d+\frac{1}{2}\sum\left(\sigma_{\alpha}^{-1}\left(1-\alpha\right)\sigma_{1}+\alpha\sigma_{2}\right)+\log\frac{\sigma_{\alpha}}{\sigma_{1}^{1-\alpha}\sigma_{2}^{\alpha}}\\
\\
\end{array}
\end{equation}
Unlike KLD in eqn (4), $\alpha$JSD further rely on skew variables
for the first and second order variables as follows \cite{nielsen2019jensen}

\begin{equation}
\begin{array}{c}
\sigma_{\alpha}=\left(\sigma_{1}^{-1}\left(1-\alpha\right)+\sigma_{2}^{-1}\alpha\right)^{-1}\\
\\
\mu_{\alpha}=\sigma_{\alpha}\left(\sigma_{1}^{-1}\mu_{1}\left(1-\alpha\right)+\sigma_{2}^{-1}\mu_{2}\alpha\right)\\
\\
\end{array}
\end{equation}
For eqn (5) and (6), we refer to $\left(\mu_{1},\sigma_{1}\right)$
and $\left(\mu_{2},\sigma_{2}\right)$ as network variable $\left(\mu,\sigma\right)$
and DPM variable $\left(\eta^{*},\tau^{*}\right)$ respectively. 

The symmetrical property of JSD can be explained using eqn (6). For
explanation purpose, we can assume the simple case of first order
variable only i.e. ignore $\left(\sigma_{1},\sigma_{2}\right)$. When
we consider the case of KLD in eqn (4), we arrive at the following
form $KLD=\frac{1}{2}\left(\mu_{1}-\mu_{2}\right)^{2}$. When we have
the network variable $\mu_{2}$ outputting a near zero, we end up
with a large value for KLD e.g. $KLD=0.5\left(\mu_{1}\right)^{2}$.
This is the main cause of asymmetry in KLD. On the other hand, we
compare the case where $\alpha JSD=\left(1-\alpha\right)\left(\mu_{\alpha}-\mu_{1}\right)^{2}+\alpha\left(\mu_{\alpha}-\mu_{2}\right)^{2}$
and $\mu_{\alpha}=\left(\mu_{1}\left(1-\alpha\right)+\mu_{2}\alpha\right)$.
Again when we assume $\mu_{2}\rightarrow0$ and $0\leq\alpha\leq1$,
we observe that $\alpha JSD\ll KLD$ as seen in Fig 2. Thus, the choice
of $\alpha$ can alleviate $\alpha JSD$ from the asymmetry problem
that plagues KLD. For a working case, when both $\mu_{1},\mu_{2}$
have similar values, both KLD and $\alpha$JSD should correctly be
computing a value close to zero as seen in Fig 2.

\begin{figure*}
\begin{centering}
\includegraphics[scale=0.35]{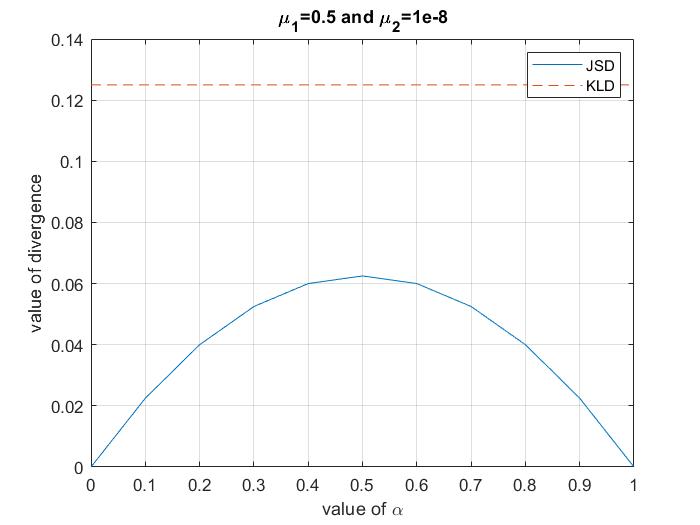}$\quad$\includegraphics[scale=0.35]{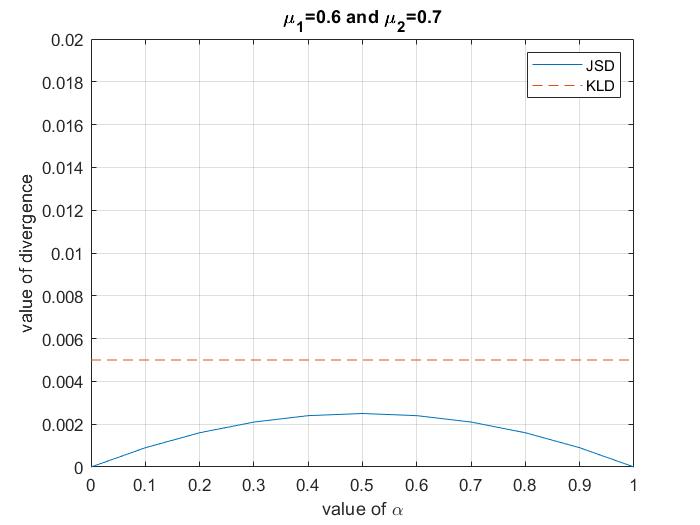}
\par\end{centering}
\caption{The asymmetry problem in KLD vs $\alpha$JSD}

\end{figure*}

\subsection{Variational inference of Dirichlet process Gaussian mixture}

In the standard GMM as shown in Fig 3, the hidden variables are updated
by the point estimates, $\hat{\tau_{k}},\hat{\eta_{k}},\hat{\varsigma_{nk}},\hat{\psi_{k}}$.
The expectation-maximization algorithm allows us to obtain point estimates
using the closed form equations \cite{bishop2006pattern,Lim2016}
below

\begin{equation}
\begin{array}{c}
\hat{\tau_{k}}=\left\{ \frac{1}{2}\sum_{n=1}^{N}\varsigma_{nk}\right\} /\\
\left\{ \frac{1}{2}\left(\sum_{n=1}^{N}(z_{n}-\eta_{k})^{2}\varsigma_{nk}\right)\right\} \\
\\
\end{array}
\end{equation}
\begin{equation}
\begin{array}{c}
\hat{\eta_{k}}=\left\{ \sum_{n=1}^{N}\varsigma_{nk}z_{n}\right\} /\left\{ \sum_{n=1}^{N}\varsigma_{nk}\right\} \\
\\
\end{array}
\end{equation}
\begin{equation}
\begin{array}{c}
\hat{\varsigma_{nk}}=\underset{k}{\arg\max}\;\left\{ \ln\psi_{k}-\frac{1}{2}\left(z_{n}-\eta_{k}\right){}^{2}\right\} \varsigma_{nk}\\
\\
\end{array}
\end{equation}
\begin{equation}
\begin{array}{c}
\hat{\psi_{k}}=\left\{ \sum_{n=1}^{N}\varsigma_{nk}\right\} /\left\{ \sum_{k=1}^{K}\sum_{n=1}^{N}\varsigma_{nk}\right\} \\
\\
\end{array}
\end{equation}

In the variational Bayesian approach, each hidden variable is modeled
as a posterior distribution. According to variational inference \cite{bishop2006pattern},
each posterior is updated by computing its posterior expectation.
The optimal posteriors of Bayesian DPM can be defined through their
expectations below in \cite{lim2018fast} as follows

\begin{equation}
\begin{array}{c}
E\left[\tau_{k}\right]'=\left\{ \frac{1}{2}\sum_{n=1}^{N}E\left[\varsigma_{nk}\right]+\left(a_{0}-1\right)\right\} /\\
\left\{ b_{0}+\frac{1}{2}\left(\sum_{n=1}^{N}(z_{n}-E\left[\eta_{k}\right])^{2}E\left[\varsigma_{nk}\right]+\lambda_{0}(E\left[\eta_{k}\right]-m{}_{0})^{2}\right)\right\} \\
\\
\end{array}
\end{equation}

\begin{equation}
\begin{array}{c}
E\left[\eta_{k}\right]'=\left\{ \sum_{n=1}^{N}E\left[\varsigma_{nk}\right]z_{n}+\lambda_{0}m_{0}\right\} /\\
\left\{ \sum_{n=1}^{N}E\left[\varsigma_{nk}\right]+\lambda_{0}\right\} \\
\\
\end{array}
\end{equation}

\begin{equation}
\begin{array}{c}
E\left[\varsigma_{nk}\right]'=\underset{\varsigma_{nk}}{\arg\max}\;\left\{ -\frac{1}{2}\left(z_{n}-E\left[\eta_{k}\right]\right){}^{2}+\ln E\left[v_{k}\right]\right.\\
\left.+\sum_{l=1}^{k-1}\ln(1-E\left[v_{l}\right])\right\} \varsigma_{nk}\\
\\
\end{array}
\end{equation}

\begin{equation}
\begin{array}{c}
E\left[v_{k}\right]'=\left\{ \sum_{n=1}^{N}E\left[\varsigma_{nk}\right]\right\} /\\
\left\{ \sum_{n=1}^{N}\sum_{j=k+1}^{K}E\left[\varsigma_{nj}\right]+\omega_{0}-1\right\} \\
\\
\end{array}
\end{equation}

The above hyperparameters $\lambda_{0},m_{0},\omega_{0},a_{0},b_{0}$
can either be found empirically or treated as noninformative priors
for convenience. Earlier we have discussed on the cluster weightage,
$v=\left\{ v_{k}\right\} _{k=1}^{T}\in\mathbb{R}{}^{Z}$ in DPM, which
is responsible for model selection, as seen in Fig 3. The truncation
level $T$ refers to the initial value on the largest cluster size
we set (e.g. $T<100$ for MIT67). In \cite{bishop2006pattern}, the
$k$ clusters which are significant will show large values in $v_{k}$
and vice versa. By observing $v_{k}$, we can discard the redundant
clusters with insignificant values. When $v$ has reach optimal training,
$T$ approaches the ground truth value of $K$. This will also affect
cluster mean $\eta_{k}$, cluster precision $\tau_{k}$ and cluster
assignment $\varsigma_{nk}$. This is known as cluster prunning. Similarly,
we can use $\psi=\left\{ \psi_{k}\right\} _{k=1}^{T}\in\mathbb{R}{}^{Z}$
for GMM and apply the cluster prunning approach. 

\begin{figure*}
\begin{centering}
\includegraphics[scale=0.6]{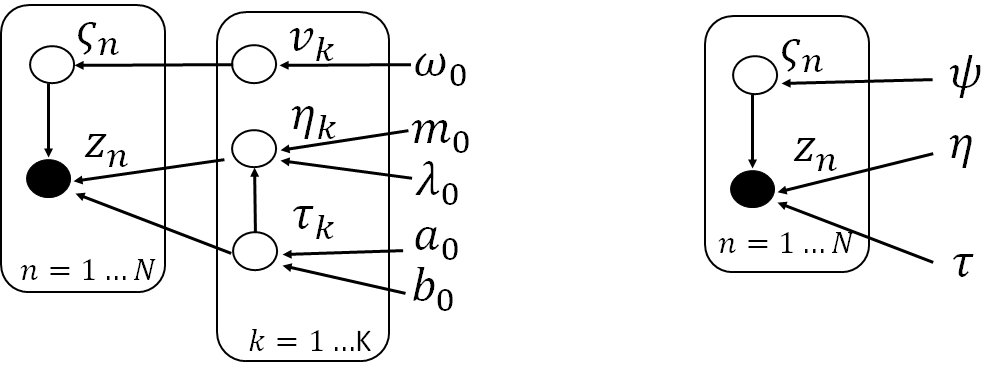}
\par\end{centering}
\caption{Graphical representation of DPM used in proposed deep model selection
(left) vs GMM used in deep clustering by other works (right).}
\end{figure*}

\section{Experiments}

\subsection{Unsupervised methods for comparison }

We compare our proposed method on image classification datasets with
the following methods.\textbf{ DPM}: Our baseline by learning DPM
directly on raw feature space, using eqn (11)-(14). \textbf{KLD}:
The method described using eqn (4) and eqn (7)-(10), which is KLD
loss \cite{lim2020} using GMM. $\alpha$\textbf{JSD}: Our proposed
method based on JSD loss using Bayesian DPM in eqn (5), (6) and eqn
(11)-(14). 

\subsection{Setup}

\textbf{Network}: We train MSE loss and $\alpha JSD$ loss separately
for the network. Starting with random weights for the encoder and
decoder, we use MSE loss to train the encoder. Then, we use $\alpha JSD$
loss to further train the encoder. We use a learning rate of 0.01
for the encoder with two hidden layers, i.e. $512-384-256-128$. Most
deep clustering methods perform all three tasks i.e. feature extraction,
dimension reduction and Type II regularizer on the same encoder's
two or three hidden layers. It is actually harder to perform deep
clustering this way as the encoder weights used for different dataset
have to be pretrained or tuned differently. Some of the best results
of these deep clustering methods require dedicated tuning. We compensate
for our lack of depth in the network using pretrained ResNet as the
feature extractor. In some datasets where the image is real, some
authors also use this strategy e.g. VaDE \cite{ji2017deep} use ResNet50
for STL-10. We only demonstrated end-to-end on MNIST, i.e. using raw
pixel. For the other datasets, we solely use ResNet18 pretrained on
ImageNet. Each image after ResNet18 feature extraction has a dimension
1x512. The $\alpha JSD$ parameter $\lambda_{3}=1$. We typically
fix $\alpha=0.65$ in eqn (6) to enable more biasing towards DPM for
the learning of the network weights. \textbf{DPM}: The hyperparameters
we use are $\omega_{0}=2000$, $a_{0}=1.25$ $b_{0}=0.25$, $m_{0}=1$
and $\lambda_{0}=0.5$. The DPM truncation levels are set at $T=200$
(CIFAR100), $T=100$ (MIT67) and at $T=50$ (CIFAR10). We also use
Kmeans for initialization.

\subsection{Experimental Results}

We refer the reader to \cite{lim2020,krizhevsky2009learning} for
more details on the datasets used in this work i.e. CIFAR10, MIT67,
and CIFAR100. We use Accuracy (ACC) for clustering as defined in \cite{CHH05}
to evaluate the performance of our method. We rerun our experiments
at least 10 rounds and take the average result in Table II.

\textbf{CIFAR10:} Most clustering papers do not directly use the raw
pixels in CIFAR10 for clustering. Thus, we use ImageNet pretrained
ResNet18 (on CIFAR10) as the image feature for DPM and $\alpha JSD$.
As a direct comparison, our baseline using DPM is able to obtain ACC
at 0.7941 while, $\alpha JSD$ is able to outperform DPM while having
more closely estimated cluster size, $\hat{K}$. For ACC, $\alpha JSD$
is slightly better than $KLD$ for CIFAR10. 

\textbf{MIT67}: MIT67 has 67 classes and lesser number of samples
than CIFAR10. It is quite useful as an intermediate dataset for deep
clustering and model selection task. However as the real images are
very complex, most deep clustering which use encoder for feature extraction
are unable to cope with this dataset. Instead, we use ImageNet pretrained
ResNet18 as the image feature. The estimated cluster size, $\hat{K}$
of $\alpha JSD$ is slightly poorer than DPM as there are much fewer
samples per class for this dataset than CIFAR10. But $\alpha JSD$
outperforms KLD in terms of ACC in this dataset. Overall $\alpha JSD$
is able to obtain the best performance.

\textbf{CIFAR100: }This is a challenging dataset as it has a fairly
large number of classes. It is difficult to find papers that target
beyond 10 or 20 classes. Furthermore, the raw pixels in CIFAR100 are
unsuitable for direct use as the object classification dataset contains
real images distorted by many image variance. Similarly, we also use
a ImageNet pretrained ResNet18 (on CIFAR100) as the image feature
so as to boost up the feature discrimination. $\alpha JSD$ is able
to consistently outperform all baselines in terms of ACC and the estimated
cluster size. Lastly, $\alpha JSD$ demonstrates that $\alpha$JSD
performs better than KLD in this dataset for ACC.

\begin{table}[h]
\caption{Experimental results}

\centering{}%
\begin{tabular}{ccccccccc}
 & \multicolumn{2}{c}{CIFAR10} &  & \multicolumn{2}{c}{MIT67} &  & \multicolumn{2}{c}{CIFAR100}\tabularnewline
\cline{2-3} \cline{3-3} \cline{5-6} \cline{6-6} \cline{8-9} \cline{9-9} 
 & ACC & $\hat{K}$ &  & ACC & $\hat{K}$ &  & ACC & $\hat{K}$\tabularnewline
DPM & 0.7941 & 17 &  & 0.6121 & 81 &  & 0.5753 & 150\tabularnewline
%$ABCD^{*},\alpha=0.65$ & \textbf{0.8845} & $\hat{K}$=12\tabularnewline
%$ABCD^{*},\alpha=0.5$ & 0.8834 & $\hat{K}$=12\tabularnewline
$KLD$ & 0.8834 & 12 &  & 0.70426 & 85 &  & 0.6145 & 118\tabularnewline
$\alpha JSD$ & \textbf{0.8845} & 12 &  & \textbf{0.71906} & 83 &  & \textbf{0.6221} & 119\tabularnewline
\end{tabular}
\end{table}

\section{Conclusion}

In this paper we discussed about deep clustering, a type of deep learning
which performs traditional clustering in deep learning feature space
using VAE. The state-of-the-art in this category is a method known
as VaDE. The problem VaDE tries to solve is how to use a KLD representation
for VAE and GMM. The problem we try to solve is how to use a JSD representation
for VAE and DPM. The novelty of using JSD is that the regularizer
is symmetric and always defined unlike KLD. The problem underlying
JSD is that no closed form solution is available. Thus, we introduced
a new variant of JSD known as the $\alpha$JSD to overcome this problem.
When using GMM for clustering, the number of clusters is required
to be defined. Instead of GMM, we proposed using DPM for $\alpha$JSD
which allows joint model selection and deep clustering. We call this
approach as deep model selection. Lastly, we compare our proposed
approach with both KLD using GMM and variational Bayes DPM on several
large class number datasets with convincing results. A possible future
work could be to exploit the neighborhood information of images for
training the latent space of VAE. This would require using different
types of regularization for VAE as discussed in Table I.

\pagebreak{}

\bibliographystyle{IEEEtran}
\addcontentsline{toc}{section}{\refname}\bibliography{allmyref}

\end{document}